\documentclass[letterpaper, 10 pt, conference]{ieeeconf}  

\IEEEoverridecommandlockouts                              

\title{\LARGE \bf
Prediction uncertainty-aware planning using deep ensembles and   constrained trajectory optimisation
}

\author{Anshul Nayak$^{1}$ and Azim Eskandarian$^{2}$
\thanks{}
\thanks{$^{1}$Anshul Nayak is with the department of Mechanical Engineering,
        Virginia Tech, Blacksburg,
         USA
        {\tt\small anshulnayak@vt.edu}}%
\thanks{$^{2}$Azim Eskandarian is the Dean of College of Engineering
and Alice T. and William H. Goodwin Jr. Endowed Chair, 
        Virginia Commonwealth University, VA 23284, USA
        {\tt\small eskandariana@vcu.edu}}}%

\newcommand{\Input}{\State \textbf{Input}}
\newcommand{\Output}{\State \textbf{Output}}

\begin{document}

\maketitle
\thispagestyle{empty}
\pagestyle{empty}

\begin{abstract}

Human motion is  stochastic and ensuring safe robot navigation in a pedestrian-rich environment requires proactive decision-making. Past research relied  on  incorporating deterministic future states of surrounding pedestrians which can be overconfident leading to unsafe robot behaviour. 
The current paper proposes a predictive uncertainty-aware planner that integrates neural network based probabilistic trajectory prediction into planning. Our method uses a deep ensemble based network for probabilistic forecasting of surrounding humans and integrates the predictive uncertainty as constraints into the planner. We compare numerous constraint satisfaction methods on the planner and evaluated its performance on real world pedestrian datasets. Further, offline robot navigation was carried out on out-of-distribution pedestrian trajectories inside a narrow corridor.

\end{abstract}

\section{Introduction}

Social navigation  remains a difficult problem as human motion is inherently stochastic and unpredictable in nature.  Ensuring safe planning of robot while not impeding the human's motion is crucial during navigation. Usually, social navigation is formulated as a collision avoidance problem where the robot finds an optimal path while avoiding humans. Prior works have focused on reactive collision avoidance methods with velocity obstacle \cite{VO}  and optimal reciprocal collision avoidance (ORCA) \cite{ORCA}. However, reactive collision  does not incorporate the predictive nature of surrounding humans while planning. Incorporating prediction of surrounding humans into planning ensures proactive decision making for the robot. Although, many approaches have integrated future predicted states of surrounding obstacles into planning, they usually considered deterministic states \cite{deterministic}. The deterministic prediction of surrounding humans can
be over confident with the planning algorithm failing to
make safe decisions as it can not account for the stochastic human motion.

The current paper addresses this problem  by proposing a prediction uncertainty-aware planning for social navigation. Our approach is modular and uses    a Bayesian neural network to forecast the predicted states as well as the associated uncertainty of  surrounding humans. We use deep ensembles \cite{Lakshminarayanan}, a popular approximate Bayesian inference method to capture predictive uncertainty.   The uncertainty-aware planner formulated as a constrained trajectory optimisation problem integrates the predictive uncertainty as constraints and  solves in a receding horizon manner  using non-linear Model Predictive control (NMPC). In particular, we compare chance constraint \cite{Blackmore} with control barrier function (CBF)\cite{Ames} based on the navigation performance of robot in a pedestrian rich environment.  Through our approach, we show how  NN based probabilistic prediction of surrounding  obstacles can be integrated into planing for safe navigation.

\section{Related Work}

Traditionally, Brito et.al \cite{Brito} proposed a model predictive contouring control (MPCC) based  collision avoidance method that compared both hard and chance constraints while avoiding humans. However, the method does not consider the future predicted states of humans while planning. In \cite{Liu},  the authors proposed a Decentralized Structural-Recurrent Neural Network (DS-RNN) which trained a policy on the reciprocal collision avoidance (ORCA) dataset while considering the future predicted states of humans  and then successfully transferred the policy from simulation to real-world for robot navigation among humans.  Similarly, Chalaki et.al \cite{Le_multi} introduced multi-robot cooperative navigation that used deep learning based human trajectory prediction using Social-LSTM and game-theoretic planning with collision avoidance  using MPC to plan. However, the prior works only considered deterministic single-step or multi-step future prediction and did not account for  uncertainty arising from stochastic human motion in planning.

 \cite{Safe_RL} presented one of the seminal works in social navigation   that enhances the safety of autonomous systems by incorporating uncertainty estimates into navigation policies. Using Monte Carlo (MC) Dropout \cite{dropout} and bootstrapping, the method provided computationally efficient uncertainty-aware collision avoidance around pedestrians.
Similarly, Zhang et. al \cite{zhang_MMP} presented a method that combines Multi-modal Motion Predictions (MMPs) using deep learning with predictive control to enable safe, collision-free navigation for mobile robots in dynamic indoor environments. Alternatively, researchers also used   Bayesian filters such as Kalman filter to predict and propagate uncertainty of surrounding obstacles into planning.  Zu et.at \cite{Hai_Zu} used extended Kalman filter (EKF) to propagate uncertainty of robot and surrounding obstacles formulating  a chance constrained trajectory optimisation for multi-agent collision avoidance. Similarly, 
\cite{Jian}  proposed a method that combines dynamic control barrier function (D-CBF)  with model predictive control (MPC) to ensure safety-critical dynamic obstacle avoidance with demonstrations through simulations and real-world experiments. However, one major drawback of these works is the use of Bayes filters to predict the future trajectory of surrounding obstacles which can be short-sighted and fail to capture long-term non-linearities.  Alternatively,   data-driven prediction methods can address such drawbacks and may be necessary for accurate incorporation of prediction into planning. 

Along similar lines, \cite{Lindemann}  presented a framework for planning in unknown dynamic environments with probabilistic safety guarantees using conformal prediction. By integrating trajectory predictions and uncertainty quantification into a Model Predictive Controller (MPC), the approach ensures provably safe navigation in  pedestrian-filled intersections within a simulated environment. 
Recently, \cite{DRCC_MPC}  introduced a distributionally robust chance-constrained model predictive control (DRCC-MPC) for safe robot navigation in human-populated environments. It uses a probabilistic risk metric to account for uncertainties in human motion, ensuring robustness against prediction errors. Our work is aligned along similar lines where we use data-driven deep ensemble based probabilistic  model to predict the trajectory of  surrounding humans \cite{DE_Anshul} and incorporate the information into planning. Also, we compare  robot navigation under different constraints such as CBF, hard constraint and chance constraints on their effectiveness for safe navigation.

\subsection*{\textbf{Contributions.}} The main contributions of the paper are:
\begin{itemize}
    \item   We present a uncertainty-aware planning algorithm that combines data-driven uncertainty-inclusive trajectory forecasting model with MPC for robot social navigation.

    \item We provide a comprehensive list of simulations comparing hard constraint, chance constraint and control barrier function for robot navigation on  publicly available pedestrian datasets such as ETH and UCY.

    \item We also validate our planning algorithm by incorporating out of distribution prediction for multiple pedestrians in a narrow corridor through offline navigation. 

\end{itemize}



\section{Uncertainty-Inclusive Trajectory Prediction}

\subsection{Probabilistic Trajectory Prediction}

The probabilistic trajectory prediction is based on the work \cite{DE_Anshul} where 
a Bayesian neural network (BNN)  is used to probabilistically predict the future state of surrounding pedestrians.

\textbf{Lemma 1.}
Usually for a  BNN,  prior distribution is assumed over weights $\mathrm{P(\theta)}$, and the posterior $\mathrm{P}(\theta|\mathrm{D})$    is learned from data $\mathrm{D} =\{X,Y\}$ using Bayes Rule. 
\begin{equation}
    {\mathrm{P}}(\theta|\mathrm{D}) = \frac{\mathrm{P}(\mathrm{D}|\theta)\,{\mathrm{P}}(\theta)}{ {\mathrm{P}}(\mathrm{D})}\,.
\end{equation}
The NN model can  predict output $y^{*}$  by marginalizing the posterior, ${p}(\theta|\mathrm{D} = {X,Y})$ over some new input data, $x^{*}$
\begin{equation}
    {p}(y^{*}|x^{*},X,Y) = { \int_{\theta}^{} {p}(y^{*}|x^{*},\theta'){p}(\theta'|X,Y) \,d\theta' }\,.
\end{equation}
$\theta'$ represents weights and  $p(\theta')$ refers to the probability of sampling from prior weight distribution.   

However, sampling from the posterior distribution  ${p}(\theta'|X,Y)$  can be computationally intractable. Therefore, approximate Bayesian inference methods like Monte Carlo (MC) dropout \cite{dropout} or deep ensembles (DE) \cite{Lakshminarayanan} can be used to predict distribution over outputs rather than deterministic prediction. In the current paper, a NN model with DE  is used to  make probabilistic forecast capturing uncertainty in human motion. 

\subsection{Deep Ensembles}\label{sec:deep_ensemble}
Deep ensembles consist of multiple independently initialized neural networks (NNs), whose outputs are averagedover the networks. Let $M$ denote the number of NNs in the ensemble, with $\mu_i(x)$ and $\sigma_i(x)$ representing the mean and variance of predictions for input $x$, where $i = 1, \ldots, M$. Balaji et.al \cite{Lakshminarayanan} showed  the ensemble of M networks as a uniformly-weighted mixture model and combined the predictions into a single Gaussian  distribution  whose mean and variance correspond to the respective mean and variance of the mixture model.

\begin{equation}
\resizebox{0.7\hsize}{!}{$
    \left \{
    \begin{aligned}
    \mu_{*}(x) &= M^{-1}\sum_{i}{} \mu_{i} (x)\,,\\
    \sigma_{*}^{2}(x) &= M^{-1}\sum_{i}{} (\sigma^{2}_{i} (x) + \mu^{2}_{i} (x)) - \mu^{2}_{*}(x)\,.
    \end{aligned}
    \right.
$}\label{predictive_var}
\end{equation}

The total predictive variance \eqref{predictive_var} can be disentangled into aleatoric uncertainty, associated with the inherent noise of the data, and epistemic uncertainty accounting for uncertainty in model  predictions. 
\begin{equation}
\resizebox{0.9\hsize}{!}{$
\begin{aligned}
     \sigma_{*}^{2}(x) &= M^{-1}\sum_{i}{} \sigma^{2}_{i}(x) &+  M^{-1}\sum_{i}\mu^{2}_{i} (x) - \mu^{2}_{*}(x)\\ 
     &=  \mathbb{E}_{i} [\sigma^{2}_{i} (x)] &+ 
     \mathbb{E}_{i} [\mu^{2}_{i}(x)] -  \mathbb{E}_{i} [\mu_{i}(x)]^{2}\\
     &= \underbrace{\mathbb{E}_{i} [\sigma^{2}_{i} (x)]}_{Aleatoric 
     \,\,uncertainty} &+ \underbrace{\mathrm{Var}_{i}[\mu_{i}(x)]}_{Epistemic \,\, uncertainty}
 \end{aligned}$}
 \label{uncertainty_disentanglement}
\end{equation}

Equation \eqref{uncertainty_disentanglement} shows that across multiple output samples, the mean of variances represents aleatoric uncertainty, while the variance of mean represents the epistemic uncertainty. We capture the total predictive uncertainty during pedestrian trajectory forecasting.

\subsection{Uncertainty Estimation}

Trajectory prediction can be considered  as regression problem for NN. Usually, standard NNs predict deterministic output say $y_{n}$ by minimizing  the MSE = $\sum_{n=1}^{N} (y_{n} - \mu({x_{n}}))^{2} $ loss but do not account for uncertainty. To capture uncertainty, we assume the outputs are sampled from a Gaussian distribution such that the final layer outputs  predicted mean $\mu({x})$ and variance $\sigma^{2}(x)$. The variance $\sigma^{2}(x)$  can be obtained using the Gaussian negative log-likelihood loss (NLL) loss.
\begin{equation} 
   -logP(y_{n}|x_{n}) = \frac{log\,\sigma^2(x_{n})}{2} + \frac{(y_{n} - \mu({x_{n}}))^{2}}{2\sigma^2(x_{n})} + C\,.
    \label{NLL_loss}
\end{equation}

$\sigma(x)$ represents the amount of noise present in the model's outputs. 

\subsection{Architecture} 
In the current research, we designed a sequence-to-sequence encoder model with attention decoder. The encoder encodes the pedestrian trajectory $\{x_{1},x_{2},\ldots,x_{T} \}$ into a latent space while the attention decoder predicts output distribution $\{\hat{y}_{T+1},\hat{y}_{T+2},\ldots,\hat{y}_{T+F} \}$ upto F steps into future. Both position and velocity $x_{t} = [\mathrm{x,y,u,v}]$ of the pedestrian are considered as inputs and the NN model is trained using Gaussian NLL loss (\ref{NLL_loss}) resulting in  a output distribution over the pedestrian states $\hat{y} \sim \mathcal{N}(\hat{\mu}, \hat{\Sigma})$ which will be incorporated during planning.


\section{Prediction Uncertainty-Aware Planning}
\subsection{Robot and Obstacle Model}

Consider the scenario where a robot shares workspace $\mathcal{W} \subseteq \mathbb{R}^{2}$ with  obstacles. We model  robot r as $p_{r}^{t_{0}}$ as an enclosing circle with radius $r_{o}$ at time $t_{0} \in \mathbb{Z}^{+} $. The non-linear dynamics of the robot which is locally Lipschitz is formulated using a discrete-time equation, 
\begin{equation}
    {x}_{t+1} = f({x}_{t}, u_{t})
\end{equation}where $x_{t} = \begin{bmatrix}
    p_{r}^t, \psi_{r}^{t}, \dot{p}_{r}^{t},\dot{\psi}_{r}^{t}, \delta_{r}^{t}
\end{bmatrix} \in \mathcal{X} \subset \mathbb{R}^{n}$ represents the state of the robot at time $t \in \mathbb{Z}^{+}$ while $u_{t} =
\begin{bmatrix}
    F_{r}^{t}, \dot{\delta}_{r}^{t}
\end{bmatrix}\in \mathcal{U} \subset \mathbb{R}^{m}$ represents the control input.

 Each pedestrian $i \in \mathcal{I}_{i} = {1,2,...,n_{i}} \subset N $ can be represented with initial  position $p_{i}^{t_{0}} \in \mathbb{R}^{2}$.  We model the pedestrian as a circle with radius $a_{0}$. To account for the predicted stochastic motion of the pedestrians,  we consider the  future states with the associated uncertainty of individual pedestrian obtained from NN. For the ith moving pedestrian, the probabilistic future state is represented as 
$\mu_{i}(t) = \begin{bmatrix}
    \mu_{x} \,\, \mu_{y} 
\end{bmatrix}^{T}$ and covariance $\Sigma_{i}(t) \in \mathbb{R}^{2}$ at step t.  
 
\subsection{Hard Collision Constraint}

\textbf{Lemma 2.} Let $\Sigma_{i}(t)$ represents the real symmetric positive semi-definite covariance matrix. Then, there exists an orthogonal matrix Q and a diagonal matrix $\Lambda$ such that
\begin{equation}
\Sigma = Q\Lambda Q^{T}
\end{equation}
where $\Lambda = diag(\Lambda_{1}, \Lambda_{2})$ with $\Lambda_{1}, \Lambda_{2} \geq 0$. Q contains the eigen vectors of the $\Sigma$ and defines the orientation of the ellipse.  The square roots of the eigenvalues $\sqrt{\Lambda_{1}}, \sqrt{\Lambda_{2}}$
represent the lengths of the semi-major and semi-minor axes of the ellipse, respectively.

Assume at any time t, the robot operates within the workspace with $n_{i}$ pedestrians. The collision condition between robot r and any pedestrian i must satisfy the following  condition. 

\begin{equation}
    C_{i}^{t} \mathrel{:=} \{ p(t) | \,\| p(t) - \hat{p}_{i}(t) \,\| \leq r_{o} + r_{th}\}
\end{equation}

$p(t) \in \mathbb{R}^{2} \subset x_{t}$ and  $\hat{p}_{i}(t) \sim \mathcal{N}(\mu_{i},\Sigma_{i})$  represent the current robot state and the probable location of ith pedestrian respectively. $r_{th}$ is the length of the major axis of the  bounding ellipse for the pedestrian.

\textbf{Remark 1.} Let $a_{0}$ be the radius of the pedestrian, then  bounding ellipse for the pedestrian $\hat{p}_{i}(t) \sim \mathcal{N}(\mu_{i}, \Sigma_{i})$ at time t has a major axis $a_{0} + \sqrt{\Lambda_{1}}$ and the minor axis $a_{0} + \sqrt{\Lambda_{2}}$

\subsection{Linear Chance Constraint}

Previously, we formulated the collision constraints as hard constraints but the predicted collision avoidance constraints can also be formulated in a probabilistic manner using chance-constraints:

\begin{equation}
    Pr(p(t) \in C_{i}^{t}) \leq \delta, \quad i \in n_{i}
\end{equation}
where $\delta$ is the probability threshold for the robot-pedestrian collision avoidance. The probabilistic constraint can be formulated as a linear chance constraint in the form $Pr(a^{T}x \leq b) \leq \delta$ where $x \in \mathbb{R}^{n}$ is a random variable and $a\in \mathbb{R}^{n}$ and $b\in{R}$
are constants.

\textbf{Lemma 3.} Given a multivariate random variable $x \sim \mathcal{N}(\mu, \Sigma)$ and probability threshold $\delta$, then 
 chance constraint can be reformulated as linear deterministic constraint 
 \begin{equation}
     Pr(a^{T}x \leq b) \leq \delta \Leftrightarrow a^{T}\mu - b \geq c \label{eq:chance}
 \end{equation}
 where $c = erf^{-1} (1 - 2\delta)\sqrt{2a^{T}\Sigma a}$ and erf(.) is the error function defined as $erf(x) = \frac{2}{\pi}\int_{0}^{x}e^{-t^{2}} dt$. Given the probability threshold $\delta$, the $erf(.)$ and its inverse can be obtained using a lookup table. Details on using chance constraint for robot-obstacle collision avoidance have been provided in prior works
 \cite{Blackmore} \cite{Xiaoxue_Zhang}. Equation \ref{eq:chance} has been deterministically reformulated as constraint for the receding horizon planner.
\[
\resizebox{0.45\textwidth}{!}{$
\kappa_i^T(t) \left( p(t) - \Pi_{\mathcal{I}_i(t)}(\hat{p}_{i}(t)) \right) \geq 
\sqrt{2 \kappa_i^T(t) \hat{\Sigma}(t) \kappa_i(t)} \, \text{erf}^{-1}(1 - 2 \delta)
$}
\]

where 

\[
\kappa_i(t) = \frac{p(t) - \Pi_{\mathcal{I}_i(t)}(\hat{p}_{i}(t))}{\|p(t) - \Pi_{\mathcal{I}_i(t)}(\hat{p}_{i}(t))\|}
\]

is the slope of the line connecting robot \( p(t) \) and ith pedestrian \( \Pi_{\mathcal{I}_i(t)}(\hat{p}_{i}(t)) \), perpendicular to the tangent plane.

\subsection{Control Barrier Function}
Alternatively, a safety filter like control barrier function (CBF) can be used for constraint satisfaction to guarantee that the system remains in a safe region. For safety-critical control,  we consider a set $\mathcal{C}$ for a continuously differentiable function $h : \mathcal{X} \subset \mathbb{R}^{n} \rightarrow \mathbb{R}$:
\begin{equation}
    \mathcal{X} = \{x \in \mathcal{X} \subset \mathbb{R}^{n} : h(x) \geq 0 \}
\end{equation}

We refer to $\mathcal{C}$ as a safe set. The function $h$ is a control barrier function if $\frac{\partial{h}}{\partial{x}} \neq 0$ for all $x \in \partial{\mathcal{C}}$ and there exists and extended class $\mathcal{K}_{\infty}$ function $\gamma$ such that h satisfies
\begin{equation}
    \exists \, \mathbf{u} \ \text{s.t.} \ \dot{h}(\mathbf{x}, \mathbf{u}) \geq -\gamma(h(\mathbf{x})), \ \gamma \in \mathcal{K}_\infty
\end{equation}

We can extend the continuous function into discrete domain as 
\begin{equation}
\Delta h(\mathbf{x}_t, \mathbf{u}_t) \geq -\gamma h(\mathbf{x}_t), \quad 0 < \gamma \leq 1, 
\end{equation}
where $
\Delta h(\mathbf{x}_t, \mathbf{u}_t) := h(\mathbf{x}_{t+1}) - h(\mathbf{x}_t).
$
Satisfying constraint (6), we have 
$
h(\mathbf{x}_{t+1}) \geq (1 - \gamma) h(\mathbf{x}_t)
$. In our problem, the D-CBF is formulated in a quadratic form as
\[
h(x_{t}) =  \,\| p(t) - \hat{p}_{i}(t) \,\| - (r_{0} + r_{th}(t))
\]

\subsection{Receding Horizon Planning}

In the previous sections, we formulate the robot-pedestrian collision avoidance  either through hard constraints,  linear chance constraints or control Barrier function. 
The above constraints will be considered as inequality constraints for   collision avoidance problem using non-linear model predictive control (NMPC).

The MPC can be formulated as a receding horizon optimization problem.

\begin{subequations}
\begin{equation}
\begin{aligned}
    J_t^*(x_t) &= \min_{u_{t:t+N-1|t}} \, p^{*}(x_{t+N|t}) + \sum_{k=0}^{N-1} q(x_k, u_k)
\end{aligned}
\label{cost_function}
\end{equation}
\begin{equation}
\begin{aligned}
    \text{s.t.} \quad x_{k+1} &= f(x_k, u_k), \, k = 0, \dots, N-1
\end{aligned}
\label{system_dyn}
\end{equation}
\begin{equation}
\begin{aligned}
    x_k &\in \mathcal{X}, \, u_k \in \mathcal{U}, \, k = 0, \dots, N-1
\end{aligned}
\label{Constraints}
\end{equation}
\begin{equation}
\begin{aligned}
    x_{t|t} &= x_t,
\end{aligned}
\label{state_init}
\end{equation}
\begin{equation}
\begin{aligned}
    x_{t+N|t} &\in \mathcal{X}_f,
\end{aligned}
\end{equation}
\begin{equation}
\begin{aligned}
    \text{Hard:} \quad g(x_k) &< 0
\end{aligned}
\label{hard}
\end{equation}
\begin{equation}
\begin{aligned}
    \text{CBF:} \quad \Delta h(x_k, u_k) &\geq -\gamma h(x_k), \, k = 0, \dots, N-1
\end{aligned}
\label{CBF}
\end{equation}
\begin{equation}
\begin{aligned}
    \text{Chance:} \quad Pr(a^{T}x \leq b) &\leq \delta, \, a \in \mathbb{R}^{n}, b \in \mathbb{R}
\end{aligned}
\label{chance}
\end{equation}
\end{subequations}

where $q(x_{k},u_{k}) \mathrel{:=} 
\|\, x_k - x_{k}^{d} \,\|_Q + \|\, u_k \,\|_R + \|\, u_{k+1} - u_k \,\|_S$ 
represents the stage cost while $p^{*}(x_{t+N|t}) \mathrel{:=} \|\, x_{t+N|t} - x^{{t+N|t}^{d}} \|_P $ represents the terminal cost (\ref{cost_function}). P, Q, R, S represent the corresponding weighting matrices. Similarly, $\mathcal{X}, \mathcal{U }, \mathcal{X}_{f}$ ensure the states, input and terminal constraint are satisfied. \ref{state_init} initializes the state of the system at time step $t \in \mathbb{Z}. ^{+}$. Finally, \ref{hard}, \ref{CBF}, \ref{chance} correspond to the inequality constraints  which are applied alternatively to the constrained optimisation formulation and compared across different performance metrics.

  \begin{algorithm} [ht]

   \caption{Prediction uncertainty-aware Planner }
    \begin{algorithmic}
\Input  \, Current robot state $x^0$ 
\Output \, Control input for robot $u^0$
\While {robot not reach goal state or no collision}
    \State $\hat{\mu}^{1:K}_i , \hat{\Sigma}^{1:K}_i   \gets \text{Predict}(p_{i},\dot{p}_{i}) \,\, \text{for} \,i \in \mathcal{I}_N$ 
    \State $x^{1:K} \gets \text{dynamics}(x^0, u^0)$ (14b)
    \State Evaluate objective function (14a)
    \State Satisfy constraint $\mathcal{X}, \mathcal{U}, \mathcal{X}_f$
    \State Evaluate Inequality constraint g
    
    \If {Chance Constraint} 
        \State $\{g\} \gets  Pr(a^{T}x \leq b) \leq \delta $
    \Else \, {Control Barrier Function}
       \State $\{g\} \gets  \Delta h(x, u) \geq -\gamma h(x)$

    \EndIf

    \State Get control sequence  $\{u\}^{1:K-1}$
    \State Output first control input $u^*$
\EndWhile
\end{algorithmic}
\end{algorithm}

\begin{figure*}[t]  
    \centering
    {

\includegraphics[width=0.23\textwidth]{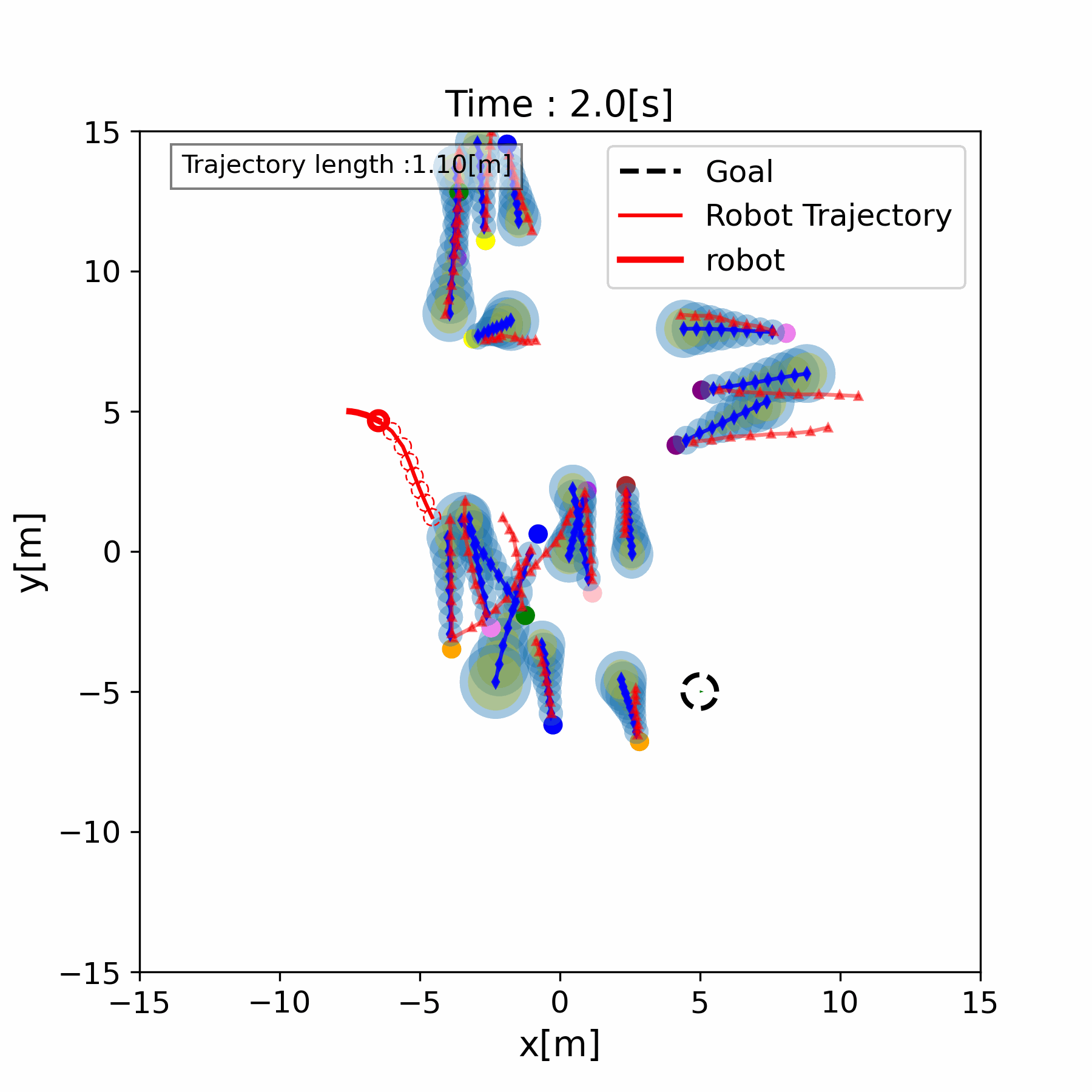}
    }
    \hfill
   {
 \includegraphics[width=0.23\textwidth]{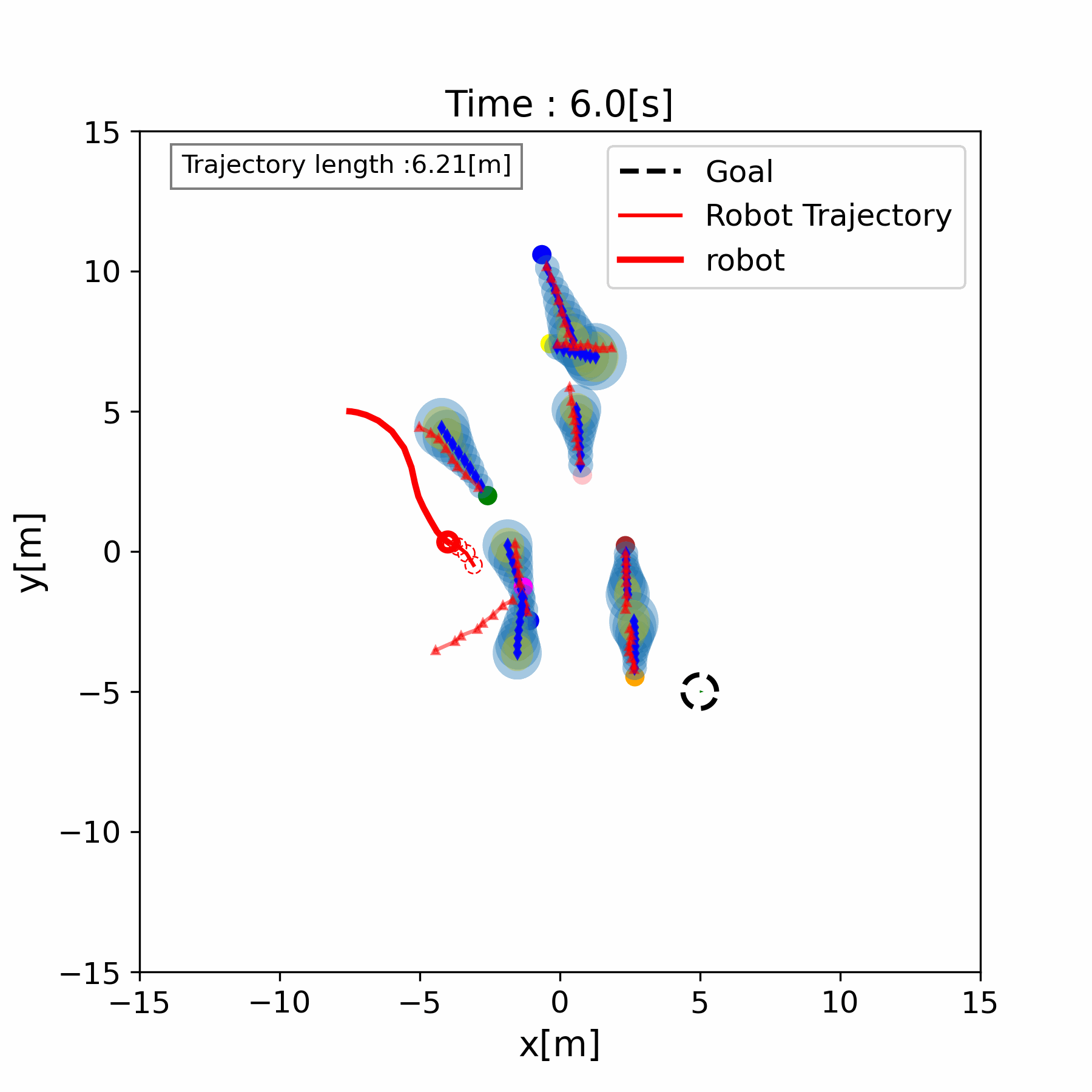}
    }
    \hfill
    {
\includegraphics[width=0.23\textwidth]{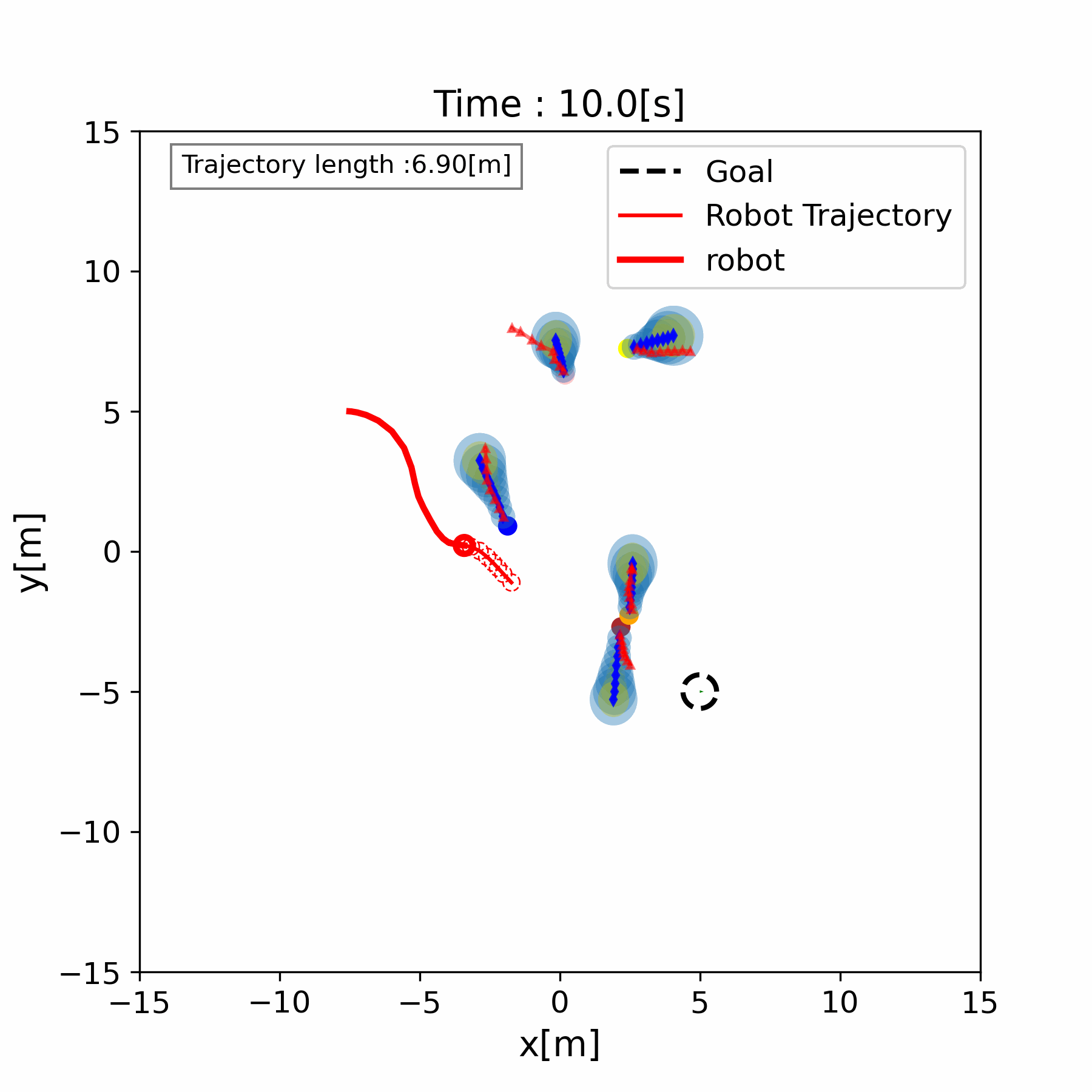}
    }
    \hfill
    {
\includegraphics[width=0.23\textwidth]{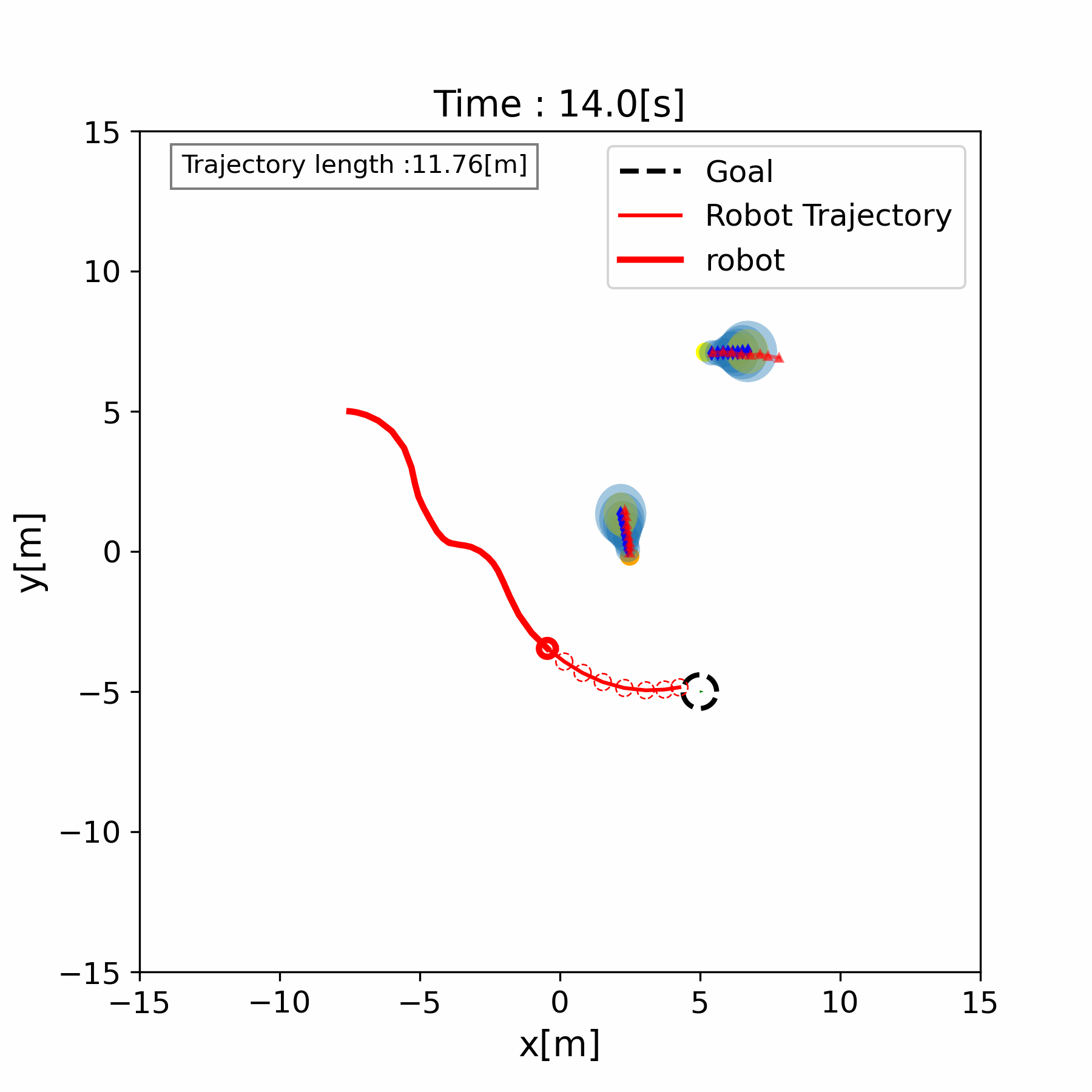}
    }
    \hfill

    \caption{Uncertainty-aware planning using MPC-CBF with a forecast horizon of 8 steps.  At t = 2 secs, the robot proactively changes its heading to avoid collision with the oncoming pedestrians. At t = 6 secs, the robot slows down as it is unable to find a feasible path to goal without hitting the pedestrian.  At t =10 secs, the robot starts to move as soon as the pedestrian crosses.  }
    \label{fig:main_figure}
\end{figure*}

\begin{table*}[h!]
\label{tab:simulation_results}
\caption{\small {Simulation result for uncertainty-aware planner comparing constraints, prediction method and forecast horizon.  Trajectory length and Total Time correspond to the length of the robot path and time it takes to navigate from start to goal if successful.  Minimum distance corresponds to the closest Euclidean distance between robot and any pedestrian. Computation time is the wall time of the solver during control update.}} 
\centering
\begin{tabular}{|c|c|c|c|c|c|c|}
\hline
\multicolumn{1}{|l|}{}            & \multicolumn{1}{l|}{} & Trajectory Length (m) & Total Time (s) & Min distance (m) & Average Computation (ms) & Success \\ \hline
\multirow{3}{*}{Constraint}       & HC                    & 38.1 +/- 0.28         & 20.04          & 0.81 +/-0.11     & 72 +/-2.9                & 1       \\ \cline{2-7} 
                                  & CBF                   & 16.8+/-0.11           & 10.8           & 1.03 +/- 0.05    & 66.5+/-2.9               & 1       \\ \cline{2-7} 
                                  & Chance                & 19.4 +/-4.5           & 21.6+/-8.05    & 2.25 +/- 0.4     & 79.8 +/- 8.7             & 0.6     \\ \hline
\multirow{2}{*}{Prediction}       & Deterministic         & 16.26                 & 17.2           & -0.14            & 51.6                     & N       \\ \cline{2-7} 
                                  & Stochastic            & 17.12                 & 20             & 0.92             & 58.4                     & Y       \\ \hline
\multirow{3}{*}{Forecast Horizon} & N = 4                 & 21.97                 & 30.4           & 2.1              & 46.7                     & N       \\ \cline{2-7} 
                                  & N = 8                 & 16.86                 & 17.2           & 1.9              & 57.3                     & Y       \\ \cline{2-7} 
                                  & N = 12                & 22.66                 & 14.4           & 2.92             & 66.7                     & Y       \\ \hline

\end{tabular}
\end{table*}

\section{Experiments}

\subsection{Implementation Details}

The current approach is modular implying that the prediction and planning are decoupled. For trajectory forecasting, a sequence-to-sequence encoder model with an attention decoder was trained using PyTorch on pedestrian dataset.  Adam optimizer with a learning rate of 8e-3 was used to compute the NLL loss for capturing predictive uncertainty. An ensemble model with M = 3 networks were  selected and each model was trained for 100 epochs with a batch size and hidden layer size equal to 64. One-tenth of the  dataset is utilised for simulation during robot navigation.

We model the robot dynamics based on the kinematic car model  assuming navigation at slow speed. The model can be represented using the non-linear  dynamics $\dot{x} = f(x,u)$.
\begin{equation}
    \begin{bmatrix}
\dot{p_{x}} \\
\dot{p_{y}} \\
\dot{\varphi} \\
\dot{v_x} \\
\dot{v_y} \\
\dot{r}\\
\dot{\delta}
\end{bmatrix}
=
\begin{bmatrix}
v_x \cos \varphi - v_y \sin \varphi \\
v_x \sin \varphi + v_y \cos \varphi \\
r \\
\frac{1}{m} F_x \\
(\dot{v_x} + \delta v_x) \frac{l_R}{l_R + l_F} \\
(\dot{v_x} + \delta v_x) \frac{1}{l_R + l_F}\\
\Delta\delta
\end{bmatrix}
\end{equation}

where $l_{r}, l_{f}$ represent the distance of COG from front and rear wheel respectively and m is the mass. The pedestrian and robot diameter are considered 0.3m and 0.6m respectively.  The control input $u_{t} =
\begin{bmatrix}
    F_{r}^{t}, \dot{\delta}_{r}^{t}
\end{bmatrix} \in \mathcal{U} \in [\pm4, \pm0.3\pi]$ correspond to the net  driving force in body direction and rate of change of steering angle respectively. The objective for the NMPC (\ref{cost_function}) represents the cost function in quadratic form. The weight matrices for  states and control inputs are $Q= [2,2,1,,1,1,1e^{-5},1e^{-5}], P =10\,Q, \, R = 0.01 \mathcal{I}_{2x2} $, $S = 100\,R$. The time step for robot control and dynamics is $\Delta t = 0.4$ secs similar to the pedestrian dataset. The optimal control problem (OCP) is formulated as a non-linear program (NLP) and solved in the CaSADi \cite{casadi} framework using IPOPT solver. The system dynamics is discretized and system state is predicted using an implicit Runge-Kutta integrator. Robot must find a feasible path to goal within a maximum simulation time of 30 seconds while  satisfying the equality and inequality constraints.

N pedestrians are randomly initialized at the start of the simulation. The robot has to navigate from an initial state $p_{0}$ to goal state $p_{g} \in \mathbb{R}^{2}$.  A run is considered successful if the robot navigates within 0.6 m of the goal and there is no collision with any human while maintaining a safe distance 0.2m. Usually, the planning horizon is considered 12 steps that corresponds to 4.8 seconds. We test our planner on  pedestrian datasets such as ETH \cite{ETH} and UCY \cite{UCY} as well as on out-of-distribution pedestrian trajectories in real-time inside a lab setup.

\subsection{Pedestrian Dataset}

Simulation results were evaluated on publicly available pedestrian datasets. 8 historical states are used to predict 12 states into future. The predicted states are normally distributed $p \sim \mathcal{N}(\mu, \Sigma)$.   We randomly select 20  pedestrian trajectories at the beginning of the simulation from the combined ETH/UCY dataset. Each pedestrian enters and exits the scene based on the duration of the trajectory within the the dataset.
Detailed results with performance metrics have been tabulated in Table I.

First, we compare the constraints namely hard  constraint, chance constraint and CBF.  We formulate the robot to navigate from an initial state to goal state while avoiding the pedestrians. The simulations were averaged over five runs under similar scenarios. Our results indicate that trajectory planning using CBF as safety constraint resulted in the smallest trajectory length with least total time. All three constraints resulted in minimum distance above the safety  threshold. Meanwhile, hard constraint follows a very conservative trajectory in order to guarantee constraint satisfaction within such a pedestrian rich environment resulting in the maximum trajectory length.

Based on the above results, we consider CBF with MPC as our uncertainty aware planner for subsequent simulations. We also tried to understand the effect of pedestrian  trajectory forecast horizon  on planning. We considered N = $\{4,8,12\}$ steps. Since the pedestrians in the dataset are tracked at 2.5 frames per second, each prediction step corresponds to 0.4 secs into future. With a very short forecast horizon of N = 4 (1.6 secs), the planner is unable to reach the goal. Meanwhile, a large forecast horizon of 4.8 secs  results in a conservative path with a large trajectory length as well as minimum distance with surrounding pedestrians. The planner with a forecast horizon of 3.2 secs i.e. N equals to 8 produces the best trajectory.

Apart from forecast horizon, prediction type of the future pedestrian trajectory whether stochastic or deterministic  also influences planned trajectory.  The current research finds out that although the inclusion of uncertainty during prediction results in a slightly conservative path with a higher total time, there are two drawbacks of using deterministic trajectory prediction over stochastic prediction  for surrounding humans during planning. First, the minimum distance is negative that results in collision. Secondly, the deterministic predictions never match the actual ground truth resulting in over confident estimate of future trajectory and planning can fail under such scenarios. 

\begin{figure*}[t]  
    \centering
    {
        \includegraphics[width=0.5\textwidth]{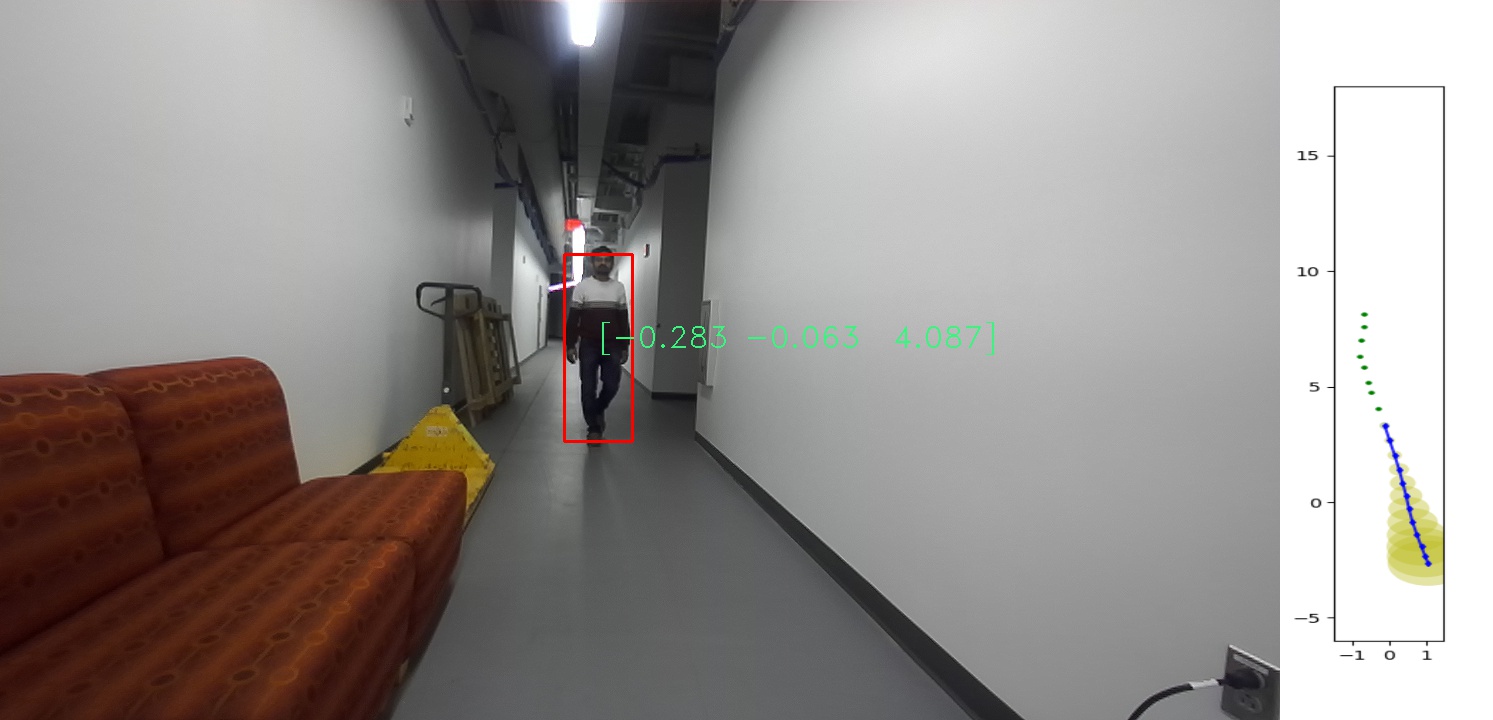}
    }
    \hfill
   {
        \includegraphics[width=0.07\textwidth]{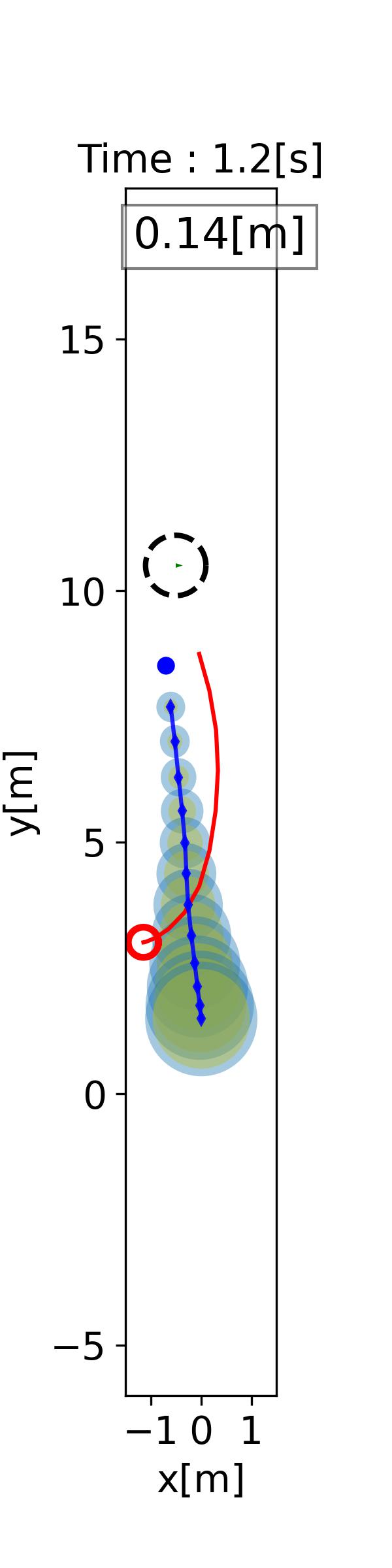}
    }
    \hfill
    {
        \includegraphics[width=0.07\textwidth]{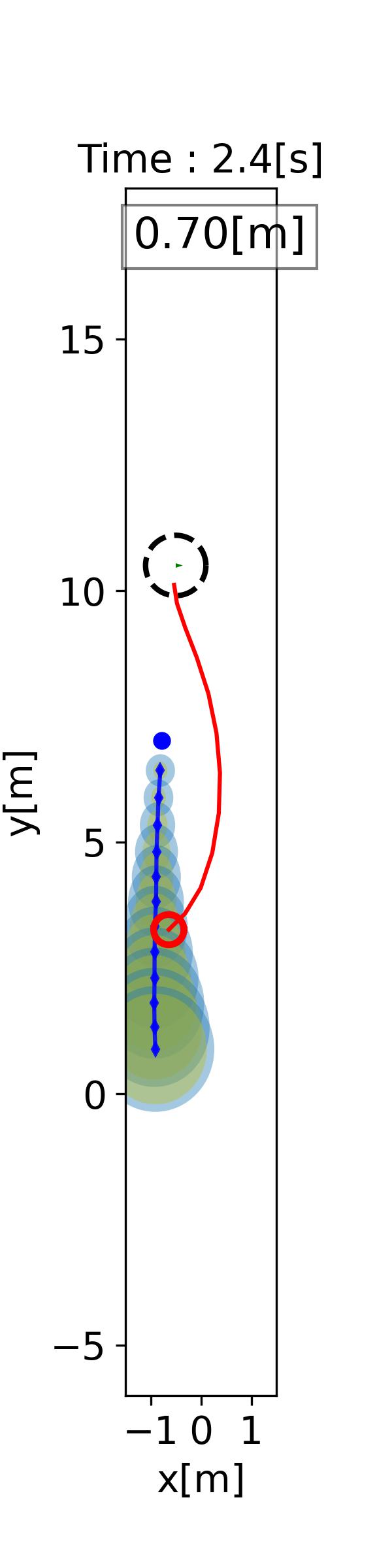}
    }
    \hfill
    {
        \includegraphics[width=0.07\textwidth]{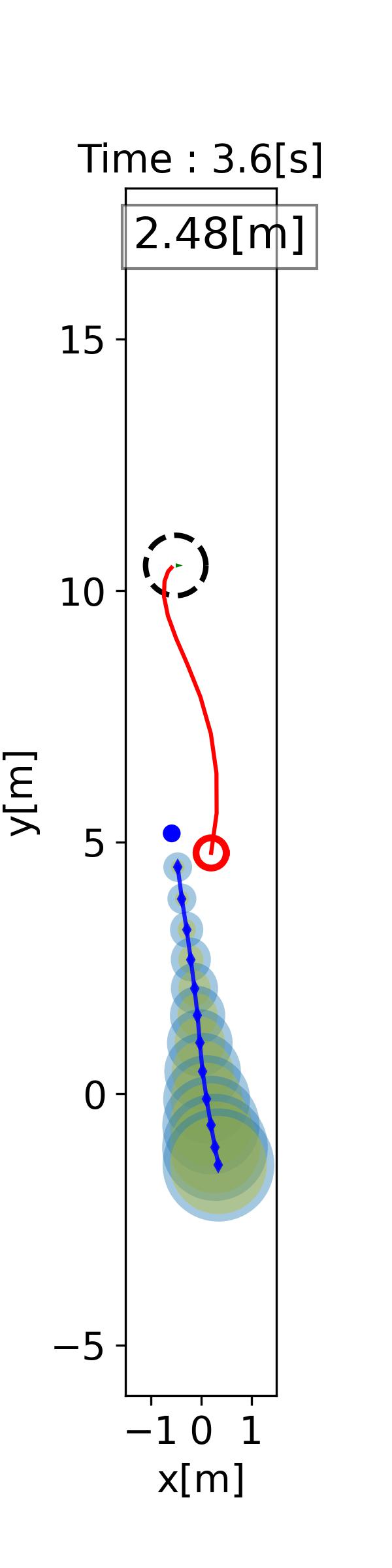}
    }
    \hfill
    {
        \includegraphics[width=0.07\textwidth]{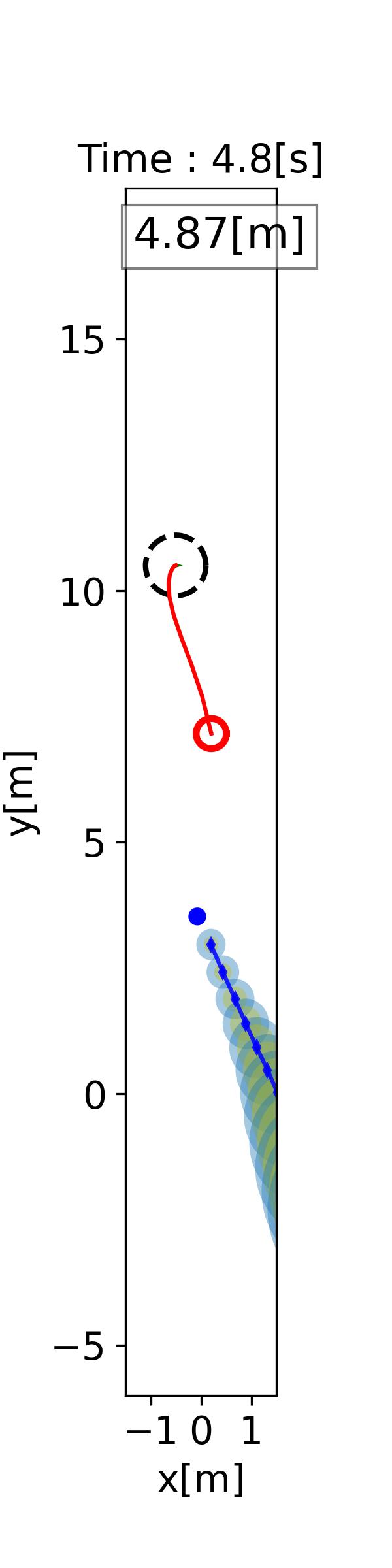}
    }
        \hfill
{
        \includegraphics[width=0.07\textwidth]{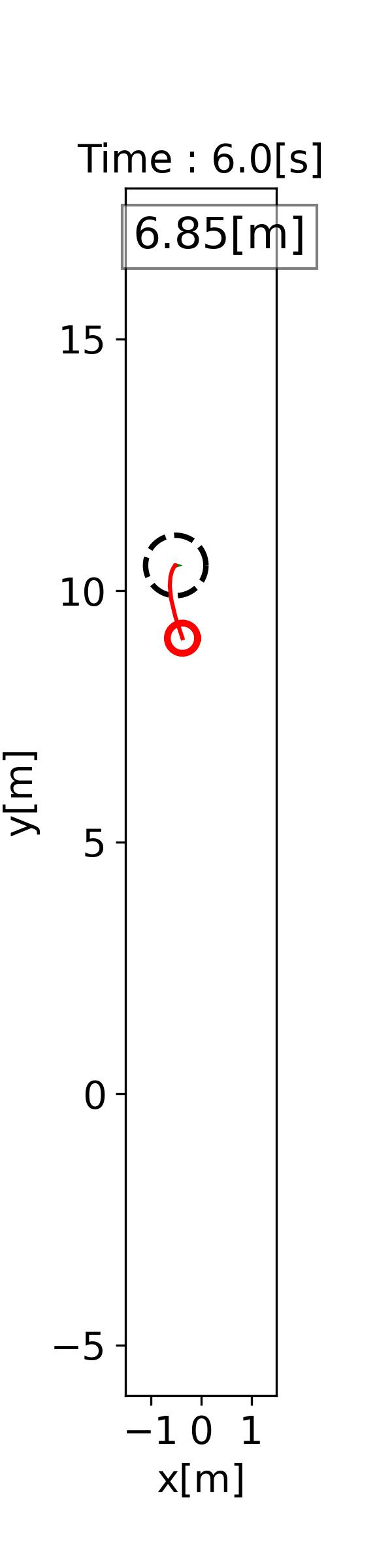}
    }
    
    \caption{Prediction uncertainty-aware offline robot navigation with a single pedestrian. At t = 2.4 secs, the robot has proactively found an optimal path to goal with a large N = 12. At t = 3.6 secs, the robot navigates ensuring a minimum distance of 0.89m. }
    \label{fig:single_pedestrian}
\end{figure*}

\begin{figure*}[h!]  
    \centering
{
        \includegraphics[width=0.5\textwidth]{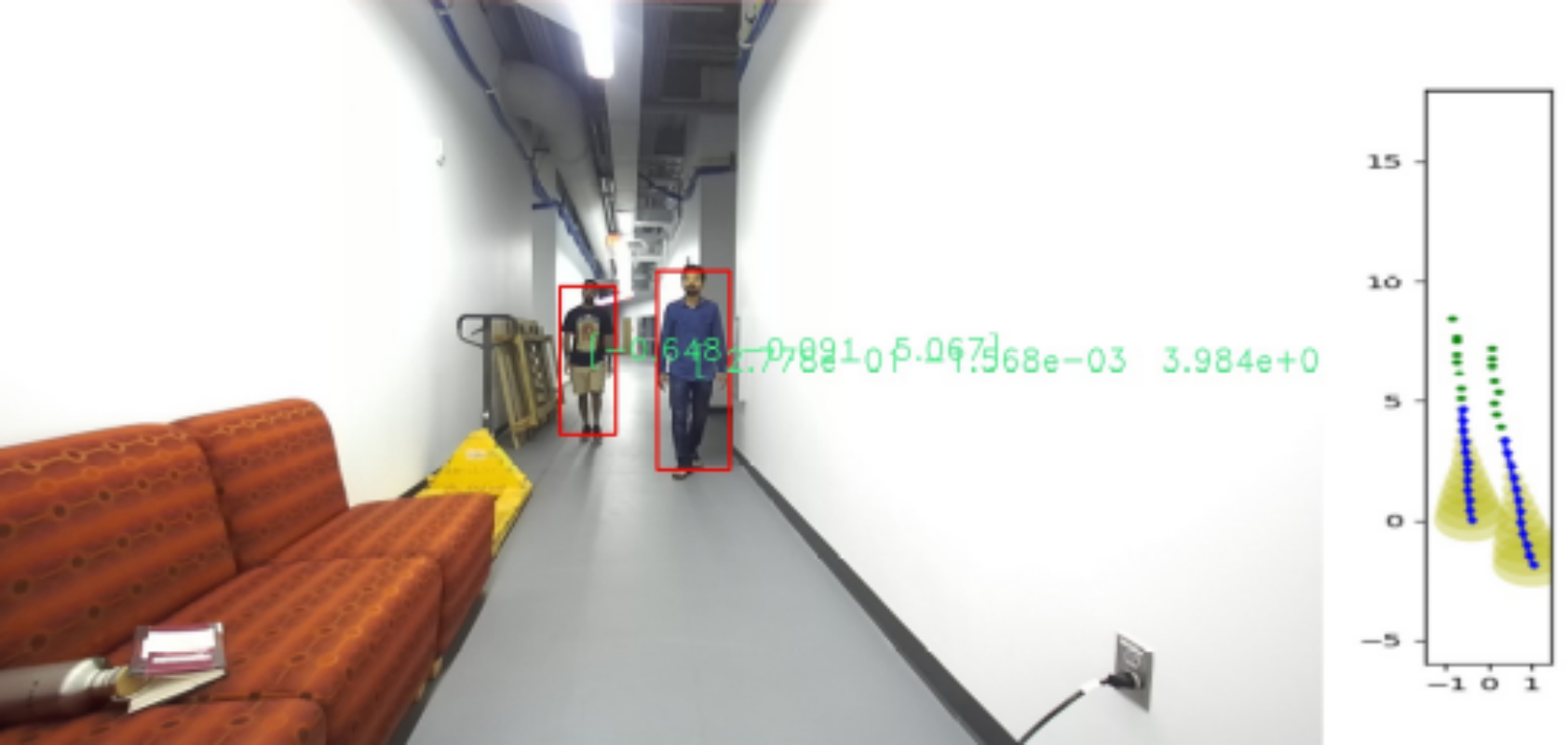}
    }
    \hfill
   {
        \includegraphics[width=0.07\textwidth]{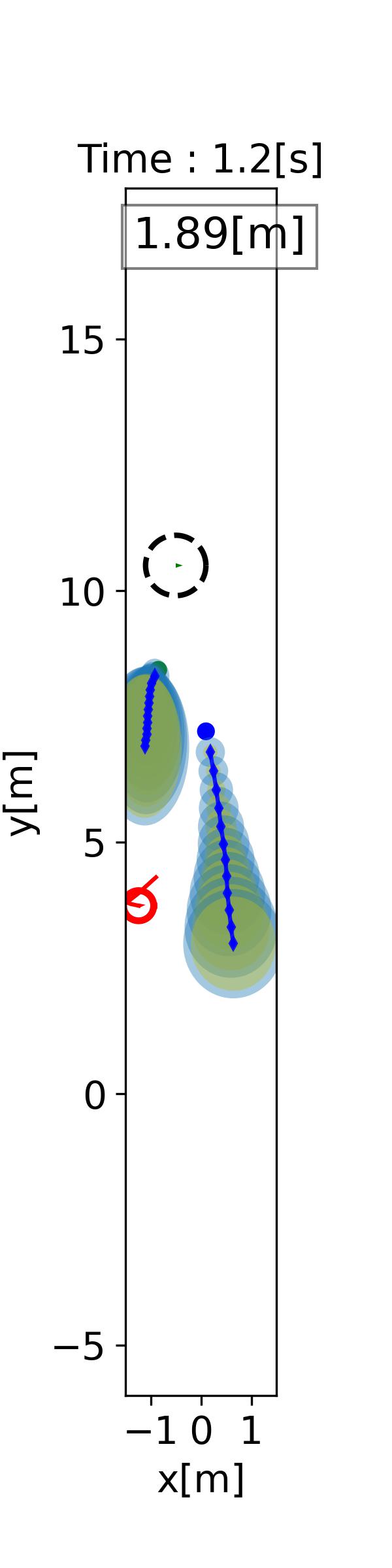}
    }
    \hfill
    {
        \includegraphics[width=0.07\textwidth]{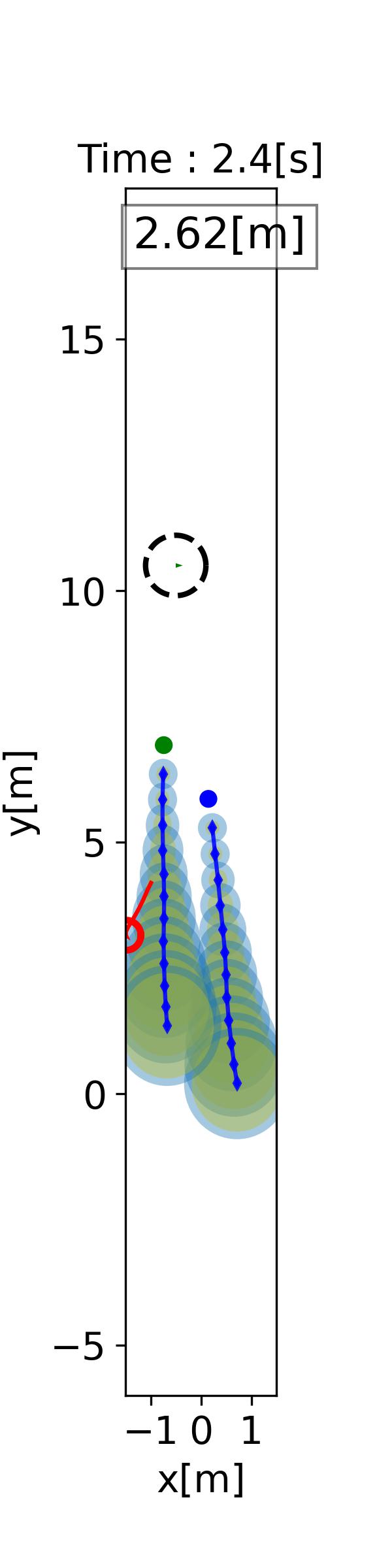}
    }
    \hfill
    {
        \includegraphics[width=0.07\textwidth]{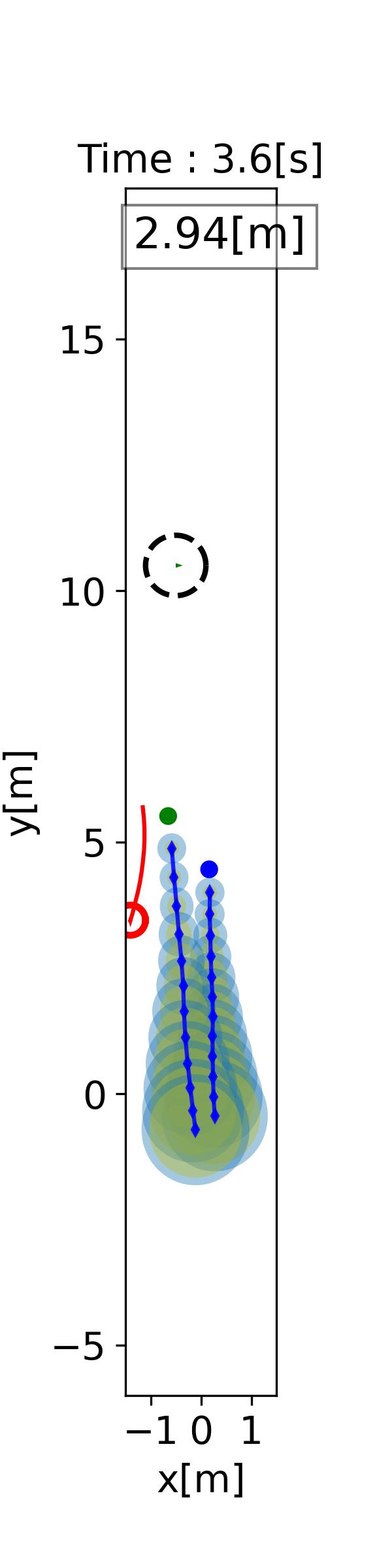}
    }
    \hfill
    {
        \includegraphics[width=0.07\textwidth]{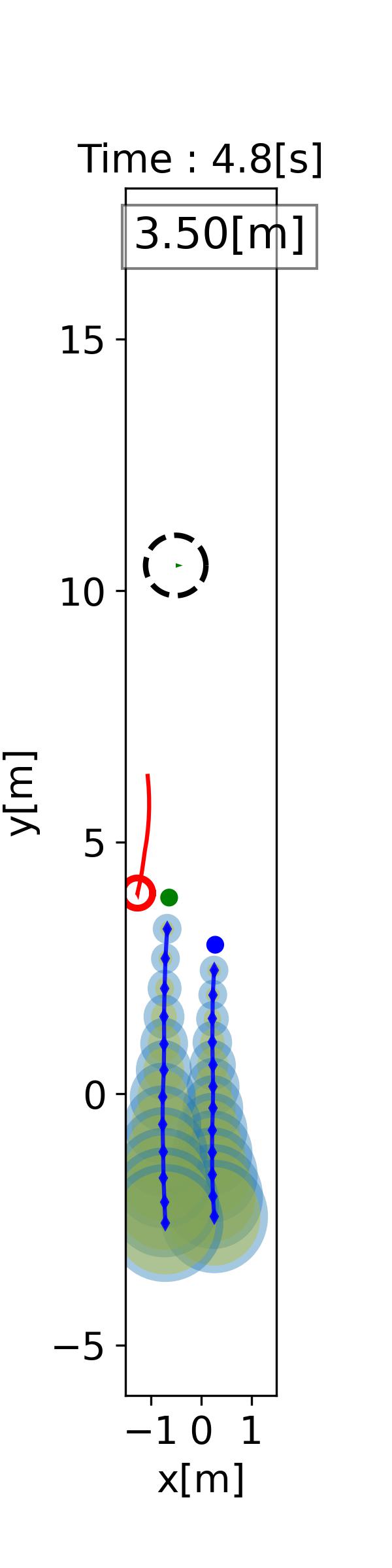}
    }
        \hfill
{
        \includegraphics[width=0.07\textwidth]{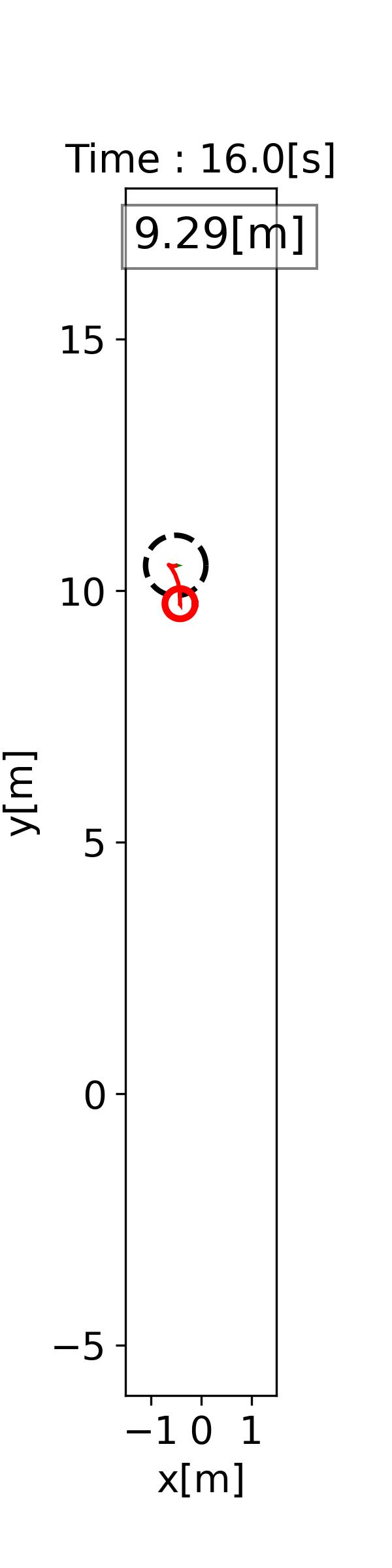}
    }
    
    \caption{Robot navigation with multiple pedestrians. Between 1.2 secs and 2.4 secs, the robot is unable to find a feasible path and even moves backward to satisfy the constraints. At t=3.6 secs, the robot manages to find a path towards goal and at t=4.8 secs, the robot avoids collision with the pedestrian while maintaining a safety margin of 0.18 m. }
    \label{fig:multiple_pedestrian}
\end{figure*}

\subsection{Out-of-Distribution Experiments}

In the previous section, we conducted planning simulations on sampled trajectories from ETH/UCY dataset. In this section, we evaluate the planner on real world out-of-distribution pedestrian trajectories. We perform real time probabilistic trajectory inference to predict the human motion in a narrow corridor and then use the trajectories for offline robot navigation. In order to collect human trajectories, we use a depth camera recording at 30 frames per second. For object detection, the camera uses Mask R-CNN \cite{He} and accurately classifies and detects all the  pedestrians inside its field of view. The sampling time is set at 12 frames such that the camera can obtain object's position and velocity every 0.4 seconds similar to the ETH/UCY dataset.
 Initially, the camera tracks and queues the pedestrians' state $\{x,y,u,v\}$ for 8 steps (3.2 secs) after which it starts to probabilistically predict the human motion in real time Figure \ref{fig:single_pedestrian}. Every single pedestrian trajectory has a duration of 8 seconds corresponding to 20 steps.  We perform two experiments with a single and multiple pedestrians navigating within a narrow corridor of width 3m.

Figure \ref{fig:single_pedestrian} shows offline robot navigation while sharing the workspace with a single pedestrian. The pedestrian walks towards the robot while the robot has to navigate to a goal state crossing the pedestrian.  Both chance and CBF constraints ensured that the robot navigates in less time while chance constraint finds a feasible path  while maintaining the smallest minimum distance with the pedestrian (Table \ref{tab:Experimental}). Similarly, Figure \ref{fig:multiple_pedestrian} shows robot navigation within a workspace with multiple pedestrians. Both pedestrians walk in the same direction with one walking ahead of the other to make the scenario challenging for the robot to navigate within such a narrow workspace. We show the result for MPC with chance constraints in Figure \ref{fig:multiple_pedestrian}.

\begin{table}[htbp]
    \centering
    \resizebox{.9\columnwidth}{!}{
    \begin{tabular}{|p{1.25cm}|p{1.25cm}|p{1.25cm}|p{1.0cm}|p{1.0cm}|p{1.0cm}|}
        \hline
        \textbf{Pedestrian} & \textbf{Constraint} & \textbf{Trajectory Length (m)} & \textbf{Total Time (s)} & \textbf{Min Distance (m)} & \textbf{Avg Time (ms)} \\ \hline
        \multirow{3}{*}{Single} & HC & 7.31 & 16.8 & 0.73 & 16.08 \\ \cline{2-6} 
         & CBF & 7.73 & 6.8 & 0.89 & 14.67 \\ \cline{2-6} 
         & Chance & 7.66 & 7.2 & 0.42 & 11.8 \\ \hline
        \multirow{3}{*}{Multiple} & HC & 14.52 & 17.2 & 0.42 & 21.05 \\ \cline{2-6} 
         & CBF & 7.5 & 6.0 & 0.28 & 15.23 \\ \cline{2-6} 
         & Chance & 9.31 & 16.4 & 0.18 & 22.53 \\ \hline
    \end{tabular}}
    \caption{Comparison of performance metrics for robot navigation  for single and multiple pedestrians across narrow corridor.}
    \label{tab:Experimental}
\end{table}

\section{Conclusions and Future Works}

The current paper proposes an uncertainty-aware planner that takes into consideration the predictive uncertainty of surrounding humans during planning ensuring proactive decision-making. The planner was evaluated on publicly available pedestrian datasets as well as real pedestrian trajectory inside a narrow corridor. Results indicate that both CBF and chance constraint outperformed the hard constraint in terms of time to goal,  trajectory length and minimum distance. Further, in closed spaces, hard constraint based planner was unable to find a feasible path or resulted in collision when uncertainty of surrounding pedestrians are considered. Looking forward, we aim to consider cooperative planning where the human also adapts to the robot behaviour.

\addtolength{\textheight}{-8cm}   






\end{document}